\documentclass[11pt,a4paper]{article}
\usepackage[hyperref]{acl2021}
\usepackage{times}
\usepackage{latexsym}

\usepackage{microtype}
\usepackage{multirow}
\usepackage{booktabs}
\usepackage{enumitem}
\usepackage[normalem]{ulem}
\useunder{\uline}{\ul}{}
\usepackage{caption}
\usepackage{subcaption}
\usepackage{graphicx}
\usepackage{verbatim}
\usepackage{ifthen}

\aclfinalcopy 
\def\aclpaperid{3461} 

\newboolean{showcomments}

\setboolean{showcomments}{true}  
\newcommand{\ta}[1]{\ifthenelse{\boolean{showcomments}}{{\color{orange}tal:[#1]}}{}}
\newcommand{\nik}[1]{\ifthenelse{\boolean{showcomments}}{{\color{cyan}nik:[#1]}}{}}
\newcommand{\sof}[1]{\ifthenelse{\boolean{showcomments}}{{\color{red}sof:[#1]}}{}}
\newcommand{\liz}[1]{\ifthenelse{\boolean{showcomments}}{{\color{blue}liz:[#1]}}{}}
\newcommand{\suchin}[1]{\ifthenelse{\boolean{showcomments}}{{\color{magenta}sg:[#1]}}{}}
\newcommand{\nas}[1]{\ifthenelse{\boolean{showcomments}}{{\color{brown}nas:[#1]}}{}}

\title{All That's `Human' Is Not Gold: \\ Evaluating Human Evaluation of Generated Text}

\author{Elizabeth Clark$^1$ \qquad Tal August$^1$ \qquad Sofia Serrano$^1$ \qquad Nikita Haduong$^1$ \\ \bf{Suchin Gururangan$^1$} \qquad \bf{Noah A. Smith$^{1,2}$} \\
$^1$Paul G. Allen School of Computer Science \& Engineering, University of Washington \\
$^2$Allen Institute for Artificial Intelligence \\
\texttt{\{eaclark7,taugust,sofias6,qu,sg01,nasmith\}@cs.washington.edu}}

\begin{document}
\maketitle
\begin{abstract}
Human evaluations are typically considered the gold standard in natural language generation, but as models' fluency improves,
how well can evaluators detect and judge machine-generated text?
We run a study assessing non-experts' ability to distinguish between human- and machine-authored text (GPT2 and GPT3) in three domains (stories, news articles, and recipes).
We find that, without training, evaluators distinguished between GPT3- and human-authored text at random chance level. 
We explore three approaches for quickly training evaluators to better identify GPT3-authored text (detailed instructions, annotated examples, and paired examples) and find that while evaluators' accuracy improved up to 55\%, it did not significantly improve across the three domains.
Given the inconsistent results across text domains and the often contradictory reasons evaluators gave for their judgments, 
we examine the role untrained human evaluations play in NLG evaluation
and provide recommendations to NLG researchers for improving human evaluations of text generated from state-of-the-art models.

\end{abstract}

\section{Introduction}
\begin{figure}[ht]
\centering
\includegraphics[width=.48\textwidth]{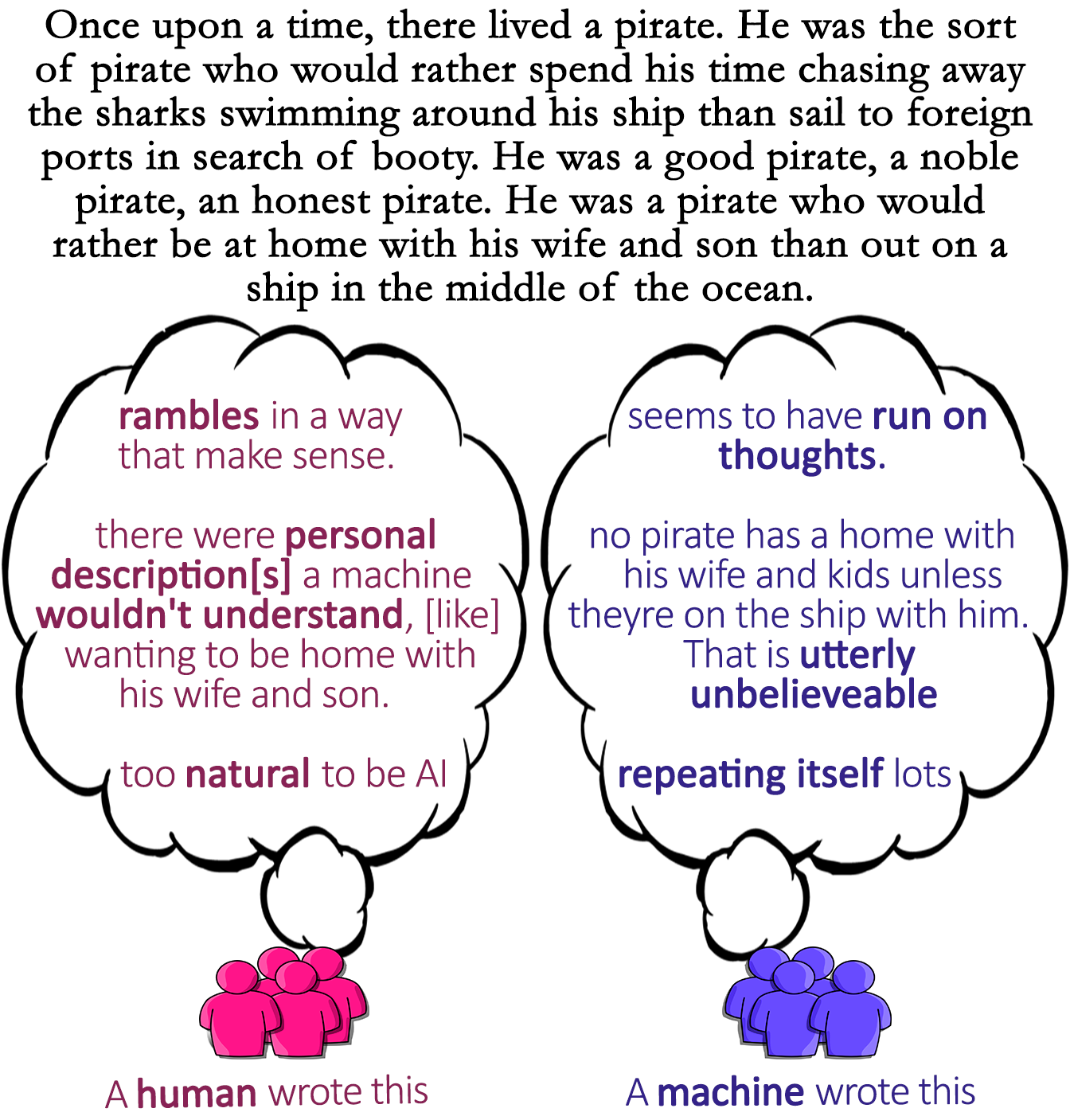} 
\caption{Excerpts from human evaluators' explanations for why they believe a GPT3-generated story (also excerpted) was written by a human (left) or a machine (right). The evaluators point to a wide range of text attributes to make their decisions, sometimes using the same aspect of the text to come to opposite conclusions.}
\label{fig:intro_fig}
\end{figure}

Human-quality text has long been a holy grail for the output of natural language generation (NLG) systems, serving as an upper bound on their performance. 
Since we lack a good way of encoding many aspects of what constitutes human-quality output in an automated method, we often must rely on human evaluation for our models.
Though evaluations with end-users in an applied setting are encouraged~\cite{belz-reiter-2006-comparing}, in practice, most human evaluations instead ask people to rate generated text's \emph{intrinsic} quality
\citep{van-der-lee-etal-2019-best, howcroft-etal-2020-twenty}.
Sometimes the generated text is explicitly compared to human-authored text \citep[e.g.,][]{liu-etal-2016-evaluate,zellers_turingadvice, Zhang2020PEGASUSPW}, but even when no human-authored text is evaluated, evaluators implicitly compare the generated text to their knowledge of language and norms within specific domains. 

Evaluators are often asked to assess a text holistically, e.g., based on its overall quality, naturalness, or humanlikeness \citep{vanderlee_journal, howcroft-etal-2020-twenty}, 
where the exact evaluation criteria is left to the discretion of the evaluator.
Though other evaluations are broken down along specific dimensions of text quality (e.g., grammaticality, coherence, etc.), \citet{novikova-etal-2017-need,novikova-etal-2018-rankme} and \citet{callison-burch-etal-2007-meta} found that these dimensions are often correlated and may be conflated in some evaluation settings.
This is concerning because, as NLG models improve, evaluators are asked to read longer passages of text conditioned on large amounts of context. In these cases, fluency-related aspects of quality (i.e., the ones that don't require careful reading of the context and meaning of the passage) are the easiest to assess, particularly in small-batch evaluations with non-expert evaluators where speed is incentivized.
This poses a challenge when collecting human evaluations for state-of-the-art language models, as errors are often content-based (e.g., factual inaccuracies or inconsistencies with the context) rather than fluency-based \citep{gpt3}, so a superficial read may not be sufficient to catch model errors.
For accurate assessments of generated text, we need human evaluations that are designed to encourage a sufficiently careful reading of the text to examine these subtler aspects of text quality.

We asked non-expert evaluators to assess the humanlikeness (operationalized as how believably human an evaluator finds a text) of text generated by current NLG models (GPT2 and GPT3) to test what current human evaluation practices can reveal about the models' quality (\S\ref{sec:exp_1}).
We found that evaluators were unable to distinguish between GPT3- and human-authored text across story, news, and recipe domains.
However, when we categorized the aspects of text the evaluators used to make their judgments, we found they primarily focused on the grammar, spelling, and style of the text.
The evaluators' responses also indicated that they underestimated the quality of text current models are capable of generating (as seen in Figure \ref{fig:intro_fig}).
To our knowledge, this paper is the first to evaluate human evaluations of GPT3-generated text across multiple domains.

We then looked at three different evaluator training methods---providing detailed instructions, annotated examples, and human-machine paired examples---to test whether we could improve evaluators' accuracy (\S\ref{sec:exp_2}). While we found including examples in the task increased the set of texts evaluators thought could be machine-generated and increased their focus on textual content, no training method significantly increased evaluators' performance consistently across domains.

Based on our results (discussed in \S\ref{sec:discussion}), we recommend moving away from small-batch evaluations with little training when collecting human evaluations of NLG models (\S\ref{sec:recommendations}).
We also encourage practitioners to consider alternative evaluation frameworks that capture the usefulness of generated text in downstream settings rather than its humanlikeness.
\section{How well can untrained evaluators identify machine-generated text?}\label{sec:exp_1}
In our first study, we ask how well untrained evaluators can distinguish between human- and machine-generated text.
This task format, inspired by the \citet{turing_test} Test, is used to compare the quality of machine-generated text to human-authored text and, as models' fluency improves, to analyze NLG models' ability to ``fool'' readers \citep{garbacea-etal-2019-judge,ippolito-etal-2020-automatic,gpt3}. 

By asking evaluators to assess the humanlikeness of the text
with only minimal instructions (see Figure \ref{fig:task}),
we observe how well untrained evaluators can detect state-of-the-art machine-generated text and which attributes evaluators focus on and think are important for detecting machine-generated text.

\subsection{The Task}
\begin{figure}[ht]
\centering
\includegraphics[scale=0.45, trim={4cm 1.7cm 4cm 2cm},clip]{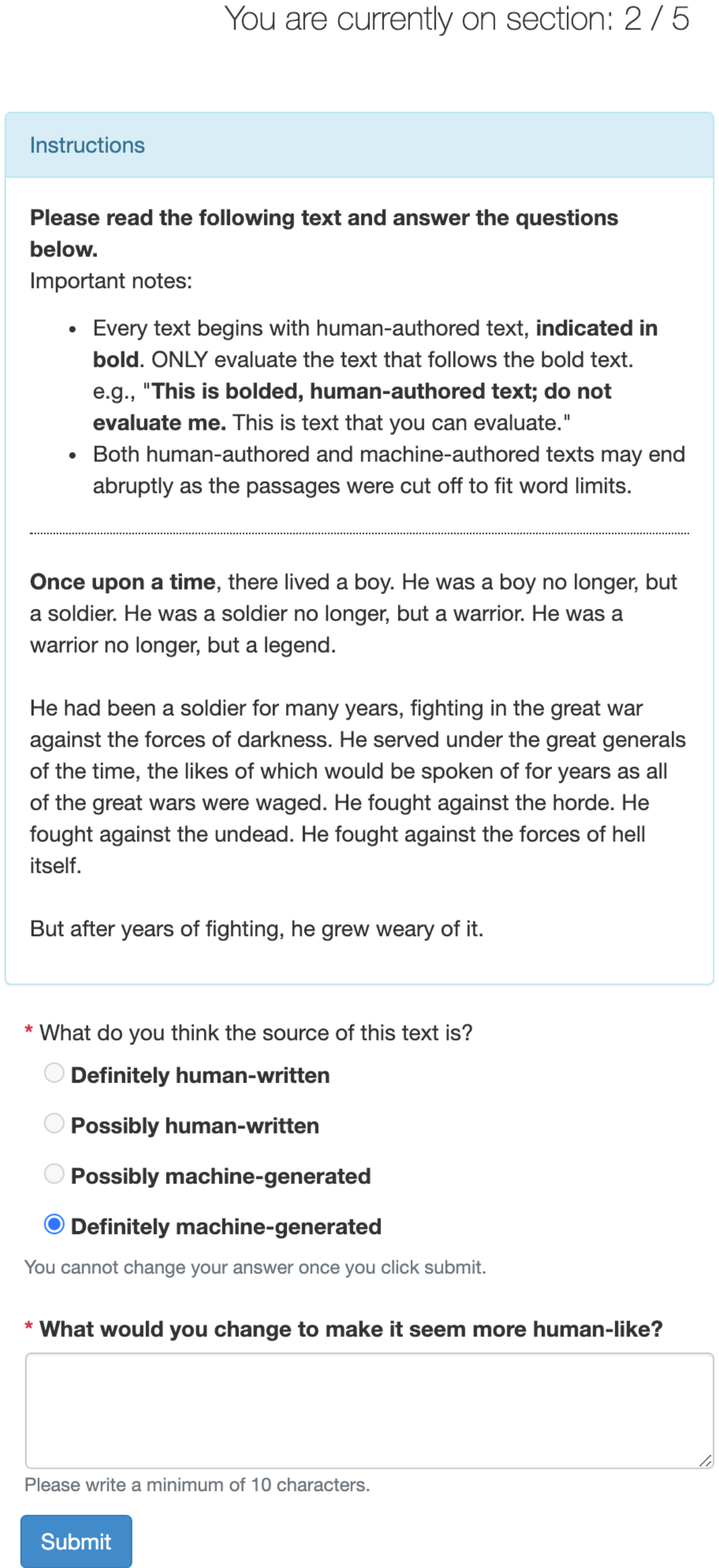}
\caption{The task interface (story domain)}
\label{fig:task}
\end{figure}

We gave evaluators 5 text passages, some of which were written by people and some generated by a model.
We asked them to rate the text on a 4-point scale \citep{ippolito-etal-2020-automatic}:
\begin{enumerate}[noitemsep]
    \item Definitely human-written
    \item Possibly human-written
    \item Possibly machine-generated
    \item Definitely machine-generated
\end{enumerate}
If they selected option 1, we asked them: ``Why did you select this rating?'' Otherwise, they were asked, ``What would you change to make it seem more human-like?''
The interface is shown in Figure \ref{fig:task}.

\subsection{Data}\label{sec:data}
We considered human- and machine-generated text in three different domains: stories, news articles, and recipes. In all three cases, we collected 50 human-authored texts in English and generated 50 texts from both the 175B parameter GPT3 model (also known as Davinci; \citealp{gpt3})\footnote{\url{beta.openai.com/}} and GPT2-XL \citep{gpt2}.\footnote{\url{huggingface.co/gpt2-xl}}
Evaluators were assigned to one domain and one model; the texts read by any given evaluator included some human-authored texts and some texts generated by their assigned model.
We only considered texts 100 words or longer,
and after reaching 100 words, all texts were truncated at the end of the next sentence.\footnote{Using NLTK; \url{www.nltk.org/}}

To generate text, we used the ``three-shot'' setting described in \citet{gpt3}, conditioning the text on three additional samples of in-domain, human-authored text, which we refer to as the \textit{priming texts} (all priming texts are in the supplementary materials and at \url{ark.cs.washington.edu/human_evals_ACL21}).
While this setting is not typically how GPT2 is used in practice, we held this approach constant to directly compare how model quality changes evaluators' ability to distinguish between texts.
For each domain, each generated text was conditioned on the same set of priming texts.
The texts were delimited with an $\langle$EOS$\rangle$ token and generated using the default GPT3 generation settings (i.e., sampling with temperature $=0.7$).

\subsubsection{Stories}
The human-authored texts came from the Reddit WritingPrompts dataset \citep{fan-etal-2018-hierarchical}.\footnote{\url{github.com/pytorch/fairseq/tree/master/examples/stories}} 
We collected all the stories that began with \textit{Once upon a time} (255 stories total) and randomly chose 50 human-authored stories from this set. For the machine-generated text, we conditioned the models on the three priming texts and on the phrase \textit{Once upon a time}.
We removed generated stories that directly copied a priming text (with $>80\%$ overlap) and regenerated those texts (9 instances with GPT2, 2 with GPT3).

This is the most open-ended of the three domains, as the story's content is virtually unrestricted, and the only creative domain.
It is also the noisiest of the human-authored datasets, as the stories were originally collected from social media comments with no quality-based filtering.

\subsubsection{News Articles}
We collected 2,111 recent local news articles from 15 different newspapers using Newspaper3k\footnote{\url{github.com/codelucas/newspaper}} (details in Appendix \ref{app:newspapers}).
After filtering out articles under 100 words, we manually filtered out articles that weren't local news or that referenced the coronavirus pandemic.
We randomly chose 50 articles to use as our human-authored news articles and another 50 to use as prompts for our generation models. 
We conditioned each generated text on the headline and first sentence from the prompt articles, along with the three priming texts.

Because the title and the first sentence of a news article often summarize its contents, the generated content must adhere to the topics they introduce. By using local, recent news, we also limit the models' ability to copy from their training data.
The models seemed to have the most trouble with this dataset structurally, e.g., generating new headlines without ending the current article or outputting invalid end-of-file tags.

\subsubsection{Recipes}
We collected 50 human-authored recipes from the RecipeNLG dataset \citep{bien-etal-2020-recipenlg}, which contains 2,231,142 recipes scraped from the web. We randomly chose an additional 50 recipes and used their titles and ingredient lists as prompts, appending them to the end of the priming texts.

This is the most closed of the three domains, as the recipe must incorporate the listed ingredients and result in the dish described by the title. Recipes are typically written in clear commands, leaving little room for surprising or unexpected text.

\subsection{Participants}\label{sec:participants}
We used Amazon Mechanical Turk (AMT) to collect the text evaluations with non-expert evaluators, commonly used in NLG evaluations \citep{van-der-lee-etal-2019-best}.
To have adequate power in our analyses (based on a power analysis with $\beta=0.8$; \citealp{card-etal-2020-little}), we had 130 different evaluators for each of the 6 task settings (3 domains $\times$ 2 models).
Each participant evaluated 5 texts each, giving us a total of 780 participants and 3,900 text evaluations.

We paid evaluators US\$1.25 for completing the task.
Following common best practice on AMT~\cite{berinsky2012evaluating}, evaluators had to have over a 95\% acceptance rate, be in the United States, and have completed over 1,000 HITs (AMT tasks).
We excluded evaluators' work if their explanations were directly copied text from the task, 
did not match their responses, did not follow the instructions, or were short, vague, or otherwise uninterpretable.
Across experiments, 445 participants (18.6\%) were rejected and not included in the \S\ref{sec:exp_1} results (780 approved participants) and \S\ref{sec:exp_2} results (1,170 approved participants).

\begin{table*}
\centering
\begin{tabular}{lrlrrrrrrr}
\toprule
Model & \begin{tabular}[c]{@{}l@{}}Overall\\ Acc.\end{tabular} & Domain & Acc. & $F_1$ & Prec. & Recall & Kripp. $\alpha$ & \begin{tabular}[c]{@{}l@{}} \% \\ human\end{tabular} & \begin{tabular}[c]{@{}l@{}} \% \\ confident\end{tabular} \\
\midrule
\multirow{3}{*}{GPT2} & \multirow{3}{*}{*0.58} & Stories & *0.62 & 0.60 & 0.64 & 0.56 & 0.10 & 55.23 & 52.00 \\
 &  & News & *0.57 & 0.52 & 0.60 & 0.47 & 0.09 & 60.46 & 51.38 \\
 &  & Recipes & 0.55 & 0.48 & 0.59 & 0.40 & 0.03 & 65.08 & 50.31 \\
 \midrule
\multirow{3}{*}{GPT3} & \multirow{3}{*}{0.50} & Stories & 0.48 & 0.40 & 0.47 & 0.36 & 0.03 & 62.15 & 47.69 \\
 &  & News & 0.51 & 0.44 & 0.54 & 0.37 & 0.05 & 65.54 & 52.46 \\
 &  & Recipes & 0.50 & 0.41 & 0.50 & 0.34 & 0.00 & 66.15 & 50.62 \\
 \bottomrule
\end{tabular}
\caption{\S\ref{sec:exp_1} results, broken down by domain and model, along with the $F_1$, precision, and recall at identifying machine-generated text, Krippendorff's $\alpha$, \% human-written guesses, and  \% confident guesses (i.e., \textit{Definitely} machine- or human-authored). * indicates the accuracies significantly better than random (two-sided $t$-test, 
for Bonferroni-corrected $p<0.00333$).}
\label{tab:exp_1_results}
\end{table*}

\subsection{Results}
Overall, evaluators choosing between human and GPT2-generated text correctly identified the author of the text 57.9\% of the time,\footnote{Unless otherwise noted, all analyses binned the responses into 2 categories (\textit{human} and \textit{machine}).} but the evaluators choosing between human- and GPT3-generated text only guessed correctly 49.9\% of the time (Table \ref{tab:exp_1_results}), compared to 50\% random chance.
While the accuracy of classifying GPT2- vs. human-authored text is significantly\footnote{$t_{388}=6.58$, $p<0.0001$} different from chance, evaluators' accuracy distinguishing GPT3- and human-authored text is not.\footnote{$t_{388}=-0.09$, $p=0.93$}
This remains the case regardless of text domain; we failed to find any evidence that evaluators' accuracy on any one domain for GPT3 differs from the overall GPT3 accuracy of $\approx50$\%.\footnote{ANOVA with $F_{2,390}=0.78$, $p=0.46$}
The story texts saw the biggest drop in evaluator accuracy from GPT2 to GPT3 (62\% to 48\%, Cohen's $d=0.57$).
The distribution of evaluators' scores are shown in Appendix \ref{app:exp1_histograms}.

In Table \ref{tab:exp_1_results}, we see other statistics worsen as well between GPT2 and GPT3: how well evaluators identified the machine-generated text ($F_1$, precision, and recall), evaluators' agreement (Krippendorff's $\alpha$, a measure of annotator agreement that corrects for the probability of random agreement), and the percent of guesses that the text was human-written (\% human).
Given that the texts are equally likely to be human- and machine-written, there are disproportionately many \textit{human} guesses, making up two thirds of the responses in the GPT3 experiments.
Despite the significantly lower scores, evaluators' confidence (the percent of \textit{Definitely} responses) remains fairly constant across conditions.

\subsection{Analysis}
Taken on its own, the evaluators' difficulty identifying GPT3-generated text compared to GPT2 points to the improvement of new NLG models.
However, it also points to concerns about extending current human evaluation methodologies to state-of-the-art text generation.
In particular, the evaluators' explanations reveal underlying confusion and misconceptions about state-of-the-art NLG.

To better understand what untrained evaluators focused on in the text to make their decisions, the authors annotated 150 random responses from the evaluators who distinguished between human- and GPT3-generated text (see Appendix \ref{app:annotation} for annotation details).
We divided the text annotation labels into three categories: \textit{form}, \textit{content}, and \textit{machine capabilities}. \textit{Form} qualities focus on the format, style, and tone of the text, while \textit{content} focuses on the text's meaning. We also coded for comments that explicitly referenced people's perceptions of what types of language machines are capable (or incapable) of generating (\textit{machine capabilities}).

We found nearly twice as many comments about the form of the text than the content (\textit{form}: 47\% of labels, \textit{content}: 25\%). Evaluators in our sample focused most on the spelling, grammar, or punctuation of the texts (45 out of 150 comments) and the style or tone of the text (24 out of 150 comments). However, these dimensions of text are unlikely to be helpful in identifying text generated by current models, considering that GPT3 has already been shown to generate fluent text and to adapt easily to new generation domains \citep{gpt3}.

We also found that the reasons evaluators gave for their answers often contradicted each other. The formality of the text, spelling and grammar errors, and clarity were all cited to justify both \textit{human} and \textit{machine} judgments.
This was also reflected in the low agreement scores between evaluators, with Krippendorff's $\alpha\approx0$ across domains.

Evaluators' expectations about what NLG models are capable of ranged from thinking their text is already indistinguishable from human-authored text (``I have no idea if a human wrote anything these days. No idea at all.'') to doubting machines' ability to use basic language (``Usually AI has terrible grammer [sic] and messes up.'').
But overall we found most evaluators' beliefs about generated language underestimated or misunderstood current NLG models, as seen in Appendix \ref{app:HUM}.
\section{Can we train evaluators to better identify machine-generated text?}\label{sec:exp_2}

Given evaluators' inability to distinguish GPT3- and human-authored text and their inconsistent reasoning for their decisions, we investigated whether there were simple ways of improving evaluators' ability to spot attributes of GPT3-generated text.
Inspired by crowdsourcing research on guiding workers on writing or other subjective tasks~\cite{kim2017mechanical, mitra_crowdsourcing}, we tested three \emph{lightweight} evaluator-training methods to see if we could improve people's ability to identify machine-generated text while maintaining the short, low-cost nature of the evaluations.

\subsection{Evaluator Training Methods}
We considered 3 evaluator trainings that can be added to the beginning of a human evaluation task, at most requiring only 3 extra samples of human- and machine-generated text. 
To test the effectiveness of each type of training, we re-ran the experiments from \S\ref{sec:exp_1}, but this time, we prepended one of three  evaluator-training methods to the evaluation task: an \emph{instruction-based} training, an \emph{example-based} training, and a \emph{comparison-based} training.
Screenshots of the training interfaces are in Appendix \ref{app:training+instructions}; the full set of training materials are in the supplementary materials and at \url{ark.cs.washington.edu/human_evals_ACL21}.

Other than the training, the task setup was identical to the GPT3-based tasks in \S\ref{sec:exp_1}.
We again ran the task on Amazon Mechanical Turk across three domains (stories, news, and recipes), using the same texts.
As each individual participant was only permitted to complete one set of evaluations, the set of evaluators who received these trainings was completely disjoint from the set of evaluators from our first study.
The participants were subject to the same restrictions described in \S\ref{sec:participants} and excluded according the same criteria; we did not use the trainings to filter out evaluators.
For each domain and training method pair, we had 130 unique evaluators complete the task, giving us 5,850 text annotations from 1,170 evaluators.

\subsubsection{Training with Instructions}
To give evaluators a better sense of which parts of the text to pay attention to, we extended the original task instructions to include dimensions of the text that could be helpful for identifying machine-generated text (repetition and factuality) and ones that could be misleading (grammar, spelling, and style).
We chose these dimensions based on previous work \citep{ippolito-etal-2020-automatic} and evaluators' comments in a pilot study (see Appendix \ref{app:pilot}). 

The Instructions training was the simplest of our 3 evaluator training methods.
It was general enough to be applied across the 3 domains but provided little information about the quality and domain of text the evaluator would be rating.
It did not increase the cost of collecting evaluations (US\$1.25 per HIT) because it does not require any extra work on the part of the evaluator, though this also made it the easiest training to ignore.
The instruction-based training is the most prescriptive of the training methods, as the researcher has to choose the dimensions they want the evaluators to focus on.

\subsubsection{Training with Examples}
Our Examples training consisted of 3 practice rounds of the actual task: given a text, guess if it is machine- or human-authored.
We collected 3 additional texts in the same manner described in \S\ref{sec:data} and wrote a short explanation of which aspects of the text hinted at its source.
After an evaluator makes their guess, the correct answer and explanation are shown.
Each domain had its own set of examples and explanations.

By showing examples, this training helps set the evaluators' expectations about the quality of the human- and machine-generated text.
We paid evaluators more for completing this task (US\$1.75 per HIT) to compensate for the extra texts they needed to read.
As with the instruction-based training, while pointing out specific text dimensions can help evaluators focus on important features, it may also restrict their search space.

\subsubsection{Training with Comparison}
In the Comparison training, we took the example passages from the Examples training and paired them with a text from the opposite source (machine or human) that began with the same prompt. 
We asked evaluators to guess which of the two texts was the machine-generated one.
We then provided the correct answer to the evaluator, along with the same explanations used in the Examples training.

This training allows evaluators to directly compare human and machine texts written from the same prompt.
It is also the most expensive training, as it required evaluators to read three more passages than the Examples training; we paid evaluators US\$2.25 per HIT.

\begin{table*}[ht]
\centering
\begin{tabular}{lrlrrrrrrr}
\toprule
Training & \begin{tabular}[c]{@{}l@{}}Overall\\ Acc.\end{tabular} & Domain & Acc. & $F_1$ & Prec. & Recall & Kripp. $\alpha$ & \begin{tabular}[c]{@{}l@{}}\%\\ human\end{tabular} & \begin{tabular}[c]{@{}l@{}}\%\\ confident\end{tabular} \\
\midrule
\multirow{3}{*}{None} & \multirow{3}{*}{0.50} & Stories & 0.48 & 0.40 & 0.47 & 0.36 & 0.03 & 62.15 & 47.69 \\
 &  & News & 0.51 & 0.44 & 0.54 & 0.37 & 0.05 & 65.54 & 52.46 \\
 &  & Recipes & 0.50 & 0.41 & 0.50 & 0.34 & 0.00 & 66.15 & 50.62 \\
 \midrule
\multirow{3}{*}{Instructions} & \multirow{3}{*}{0.52} & Stories & 0.50 & 0.45 & 0.49 & 0.42 & 0.11 & 57.69 & 45.54 \\
 &  & News & 0.56 & 0.48 & 0.55 & 0.43 & 0.05 & 62.77 & 52.15 \\
 &  & Recipes & 0.50 & 0.41 & 0.52 & 0.33 & 0.07 & 67.69 & 49.85 \\
 \midrule
\multirow{3}{*}{Examples} & \multirow{3}{*}{*0.55} & Stories & 0.57 & 0.55 & 0.58 & 0.53 & 0.06 & 53.69 & 64.31 \\
 &  & News & 0.53 & 0.48 & 0.52 & 0.45 & 0.05 & 58.00 & 65.69 \\
 &  & Recipes & 0.56 & 0.56 & 0.61 & 0.51 & 0.06 & 55.23 & 64.00 \\
 \midrule
\multirow{3}{*}{Comparison} & \multirow{3}{*}{0.53} & Stories & 0.56 & 0.56 & 0.55 & 0.57 & 0.07 & 48.46 & 56.62 \\
 &  & News & 0.52 & 0.51 & 0.53 & 0.48 & 0.08 & 53.85 & 50.31 \\
 &  & Recipes & 0.51 & 0.49 & 0.52 & 0.46 & 0.06 & 54.31 & 53.54 \\
 \bottomrule
\end{tabular}
\caption{\S\ref{sec:exp_2} results, broken down by domain and training method, along with the $F_1$, precision, and recall at identifying machine-generated text, Krippendorff's $\alpha$, \% human-written guesses, and \% confident guesses (i.e., \textit{Definitely} machine- or human-authored). ``None'' training refers to the GPT3 results from \S\ref{sec:exp_1}. * indicates accuracies significantly better than None (no training; two-sided $t$-test, for Bonferroni-corrected $p<0.00333$).}
\label{tab:exp_2_results}
\end{table*}

\subsection{Results}
We found that while all 3 training methods improved evaluators' accuracy at identifying machine- vs. human-authored text over the no-training accuracy, the Examples training was the only one that showed significant improvement (see Table \ref{tab:exp_2_results}).\footnote{Tukey's HSD adjusted $p<0.003$ for distinguishing between the Examples training and no training, $d=0.25$}

Breaking down the results by domain, however, we find the Examples accuracy did not significantly increase over the no-training accuracy when considering any of the three domains individually.
Even so, the significant difference in overall performance is mainly contributed by the story domain;
when comparing evaluators' performance with no training to its Examples training counterpart, we see a change of 0.019 and 0.062 mean accuracy in the news and recipe domains, respectively, versus 0.086 on the story domain. 
This is perhaps due to the examples helping override the preconception that machines cannot generate ``creative'' text.

Across all 3 domains, the Examples and Comparison trainings produced the highest recall and $F_1$ scores for evaluators' judgments
and decreased the percentage of texts they guessed were human-written, which indicate that evaluators were willing to consider a broader set of texts to be machine-generated than the evaluators in \S\ref{sec:exp_1}. However, despite the trainings and the increased proportion of confident responses, evaluator agreement remained low across domain and training settings ($\alpha \leq 0.11)$, and higher agreement did not correspond to higher accuracy.

\subsection{Analysis}

We again annotated 150 comments along the dimensions listed in Appendix \ref{app:annotation}, divided into \textit{form}, \textit{content}, and \textit{machine capabilities} categories, this time from evaluators who received the best-performing Examples training.
As shown in Table \ref{tab:annotation}, we found that the proportion of \textit{form} comments dropped in the sample of evaluators who went through the Examples training, while the proportion of \textit{content} comments doubled. 
We also saw a drop in the number of comments mentioning evaluators' expectations of machine-generated text. While this change in focus doesn't necessarily correspond to correct judgments, \textit{content} reasons are more in-line with current NLG model capabilities~\cite{gpt3}.

\begin{table}
\centering
\begin{tabular}{llll}
\toprule
Training & \multicolumn{1}{l}{Form} & Content & \begin{tabular}[c]{@{}l@{}}Machine\\ capabilities\end{tabular} \\
\midrule
None & 47.1 & 24.6 & 28.3 \\
Examples & 32.5 & 50.0 & 17.5 \\
\bottomrule
\end{tabular}
\caption{\% of annotation labels that reference the text's form and content and the evaluator's perception of machines' capabilities}
\label{tab:annotation}
\end{table}
\section{Discussion}\label{sec:discussion}
Overall, none of our three training methods significantly improved evaluators' ability to detect machine-generated text reliably across text domains while still maintaining the small-batch nature of Amazon Mechanical Turk.
This speaks to the improving quality of NLG models, but we also found that untrained evaluators mainly focused on the format of the text, deciding if it was human or machine-generated based on whether the text was grammatically or stylistically correct. This, combined with the high percentage of \textit{human} guesses, the low recall scores for the \textit{machine} guesses,
and the evaluators' comments on their expectations of NLG models, indicates a systematic underestimation by the evaluators of the quality of machine-generated text. 
Evaluators who were trained with examples had higher expectations of machine-generated text and focused more on the text's content; however, the training was not sufficient to significantly raise evaluators' scores across all three domains.

Many of the explanations given by evaluators included references to the text that reflected human attributes or intent that they suspected machines could not generate (e.g., ``personal description a machine wouldn't understand, [like a pirate] wanting to be home with his wife and son'' from Figure \ref{fig:intro_fig} and the examples in Appendix \ref{app:HUM}).
However, current NLG models are capable of generating text with at least superficial reference to human attributes or intent, as seen in the generated story in Figure \ref{fig:intro_fig}. This assumption that machines can't generate text with these aspects of humanlikeness led many evaluators astray, and we suspect it is one cause of the low accuracy we found. 

Crowdsourcing studies dealing only with human-authored texts often include extensive training, quality checks, or coordination~\cite{10.1145/1460563.1460572, kim2017mechanical, bernstein2010soylent}.
NLG evaluations usually forego such structures, based, we suspect, on the assumption that evaluating machine-generated text requires only fluency in the language the text is generated in. 
Our results suggest otherwise. Evaluators often mistook machine-generated text as human, citing superficial textual features that machine generation has surpassed~\cite{gpt3}. 
One potential remedy for this is to focus evaluator training on debunking this misconception. We did see evidence that the increase in accuracy we saw with our Examples training was associated with fewer explanations mistakenly referencing machine capabilities, even though the training did not specifically focus on this.
\section{Recommendations}\label{sec:recommendations}
Based on our findings, if NLG researchers must run human evaluations as small-batch evaluations on Amazon Mechanical Turk or similar platforms, we recommend they train evaluators with examples.
This will help calibrate the evaluators' expectations of generated text and indicate the careful reading they may need to do to properly assess the text's quality.
Our experiments also indicate the importance of confirming with evaluators why they have made the decisions they have, as the criteria they might implicitly be evaluating may be mismatched with researchers' intended criteria.
However, other evaluation setups may be more successful on Amazon Mechanical Turk, such as long-term evaluations with qualified evaluators who have gone through an extended training (like those in \citealp{10.1145/1460563.1460572}; \citealp{zellers-etal-2019-hellaswag}) or third-party evaluator quality tools (e.g., Positly, used by \citealp{gpt3}).

However, given the increasing length of text NLG models can handle and the careful reading needed to detect many errors in generated text, we encourage NLG researchers to move away from standalone, intrinsic human evaluation tasks.
We found that, by default,
our evaluators in this evaluation setting were most likely to focus on surface-level, fluency-related aspects of quality.
We join past work \citep{belz-reiter-2006-comparing, vanderlee_journal} in recommending a move towards evaluation settings where evaluators are better motivated to carefully consider the content and usefulness of generated text.
For example, TuringAdvice \citep{zellers_turingadvice} asks evaluators to rate NLG models by their ability to generate helpful advice, and RoFT \citep{dugan-etal-2020-roft} engages evaluators through a guessing game to determine the boundary between human- and machine-generated text. Other evaluation methods ask the evaluators to directly interact with the generated text; for example, Choose Your Own Adventure \citep{clark-smith-2021-choose} and Storium \citep{akoury-etal-2020-storium} evaluate story generation models by having people write stories with the help of generated text.\footnote{Note that we initially tried a fourth training condition along these lines, where we asked evaluators to directly interact with the generated text by rewriting it to be more humanlike. We found we were unable to successfully recruit evaluators to complete this task. The rate of retention was less than 30\%, and the rejection rate was over 50\%. We found AMT was not a good platform for this type of task, at least not for the format and the price point we explored in this work.}
We see that GPT3 can successfully mimic human-authored text across several domains, renewing the importance of evaluations that push beyond surface-level notions of quality and consider whether a text is helpful in a downstream setting or has attributes that people would want from machine-generated text.

Finally, given the mixed effect we found different trainings can have on evaluators' performance and the lack of human evaluation details typically presented in NLG papers \citep{van-der-lee-etal-2019-best, howcroft-etal-2020-twenty}, we encourage NLG researchers to include details of any instructions and training they gave evaluators in their publications.
This, along with efforts to standardize human evaluation design \citep{belz-etal-2020-disentangling, howcroft-etal-2020-twenty} and deployment \citep{genie, gem}, will support future development of evaluator training procedures and the comparison of human evaluation results in future NLG evaluation work.
\section{Related Work}
A subfield of NLG analyzes the role of human evaluations, including discussions of the tradeoffs of human and automatic evaluations \citep{belz-reiter-2006-comparing, hashimoto-etal-2019-unifying}.
There are critiques and recommendations for different aspects of human evaluations, like the evaluation design \citep{novikova-etal-2018-rankme, santhanam-shaikh-2019-towards}, question framing \citep{schoch-etal-2020-problem}, and evaluation measures like agreement \citep{amidei-etal-2018-rethinking}, as well as analyses of past NLG papers' human evaluations \citep{vanderlee_journal, howcroft-etal-2020-twenty}.
Additionally, crowdsourcing literature has work on effectively using platforms like Amazon Mechanical Turk \citep[e.g.,][]{florian_crowdsourcing,oppenheimer_crowdsourcing,weld_crowdsourcing,mitra_crowdsourcing}.
In this work, we focus on the role evaluator training can play for producing better accuracy at distinguishing human- and machine-generated text, though other quality control methods are worth exploring.

Previous work has asked evaluators to distinguish between human- and machine-authored text. For example, \citet{ippolito-etal-2020-automatic} found that trained evaluators were able to detect open-ended GPT2-L-generated text 71.4\% of the time, \citet{garbacea-etal-2019-judge} reported that individual evaluators guessed correctly 66.6\% of the time when evaluating product reviews, and \citet{gpt3} found evaluators could guess GPT3-davinci-generated news articles' source with 52\% accuracy, though these results are not directly comparable to ours due to differences in the evaluation setup, data, and participants.

Finally, our findings that untrained evaluators are not well equipped to detect machine-generated text point to the importance of researching the safe deployment of NLG systems. \citet{gehrmann-etal-2019-gltr} proposed visualization techniques to help readers detect generated text, and work like \citet{grover_zellers}, \citet{ippolito-etal-2020-automatic}, and \citet{uchendu-etal-2020-authorship} investigated large language models' ability to detect generated text.
\section{Conclusion}
We found that untrained evaluators were unable to distinguish between human- and GPT3-generated text from three domains. However, we also found that the evaluators focused on surface-level text qualities to make these decisions and underestimated current NLG models' capabilities.
We experimented with three methods for training evaluators, and while example-based trainings led to increases in recall and the amount of content-based evaluations, they did not lead to significant improvements in accuracy across all domains.
Given that evaluators struggled to distinguish between human- and machine-generated text in this setting, we should shift how we think about collecting human evaluations for current NLG models.

\section*{Acknowledgments}
This research was supported in part by the Office of Naval Research under the MURI grant N00014-18-1-2670. The authors would like to thank OpenAI, specifically Bianca Martin and Miles Brundage, for providing access to GPT3 through the OpenAI API Academic Access Program. The authors would also like to thank Katharina Reinecke, the members of the CSE 599 crowdsourcing class, and the ARK group for their feedback, the reviewers for their helpful comments, and the participants who took part in our study.

\paragraph{Ethical considerations}
All experiments in this paper were approved by our institution's internal review board.
Evaluators' responses were collected and stored anonymously.
Evaluators were paid based on an estimated US\$10 per hour rate; we raised the price of the task in proportion to the added difficulty of our 3 training methods.
For each dataset we considered, its source and language are included, along with any other details we believed would be relevant to evaluators' ability to read and understand the text.
Evaluators were warned about possible risks before starting the task, namely that NLG models can generate text with harmful language or themes, and were able to leave comments about their experience at the end of the study.

\bibliographystyle{acl_natbib}
\bibliography{anthology,non_acl}

\begin{thebibliography}{40}
\expandafter\ifx\csname natexlab\endcsname\relax\def\natexlab#1{#1}\fi

\bibitem[{Akoury et~al.(2020)Akoury, Wang, Whiting, Hood, Peng, and
  Iyyer}]{akoury-etal-2020-storium}
Nader Akoury, Shufan Wang, Josh Whiting, Stephen Hood, Nanyun Peng, and Mohit
  Iyyer. 2020.
\newblock \href {https://doi.org/10.18653/v1/2020.emnlp-main.525} {{STORIUM}:
  {A} {D}ataset and {E}valuation {P}latform for {M}achine-in-the-{L}oop {S}tory
  {G}eneration}.
\newblock In \emph{Proceedings of the 2020 Conference on Empirical Methods in
  Natural Language Processing (EMNLP)}, pages 6470--6484, Online. Association
  for Computational Linguistics.

\bibitem[{Amidei et~al.(2018)Amidei, Piwek, and
  Willis}]{amidei-etal-2018-rethinking}
Jacopo Amidei, Paul Piwek, and Alistair Willis. 2018.
\newblock \href {https://www.aclweb.org/anthology/C18-1281} {Rethinking the
  agreement in human evaluation tasks}.
\newblock In \emph{Proceedings of the 27th International Conference on
  Computational Linguistics}, pages 3318--3329, Santa Fe, New Mexico, USA.
  Association for Computational Linguistics.

\bibitem[{Belz and Reiter(2006)}]{belz-reiter-2006-comparing}
Anja Belz and Ehud Reiter. 2006.
\newblock \href {https://www.aclweb.org/anthology/E06-1040} {Comparing
  automatic and human evaluation of {NLG} systems}.
\newblock In \emph{11th Conference of the {E}uropean Chapter of the Association
  for Computational Linguistics}, Trento, Italy. Association for Computational
  Linguistics.

\bibitem[{Belz et~al.(2020)Belz, Mille, and
  Howcroft}]{belz-etal-2020-disentangling}
Anya Belz, Simon Mille, and David~M. Howcroft. 2020.
\newblock \href {https://www.aclweb.org/anthology/2020.inlg-1.24}
  {Disentangling the properties of human evaluation methods: A classification
  system to support comparability, meta-evaluation and reproducibility
  testing}.
\newblock In \emph{Proceedings of the 13th International Conference on Natural
  Language Generation}, pages 183--194, Dublin, Ireland. Association for
  Computational Linguistics.

\bibitem[{Berinsky et~al.(2012)Berinsky, Huber, and
  Lenz}]{berinsky2012evaluating}
Adam~J. Berinsky, Gregory~A. Huber, and Gabriel~S. Lenz. 2012.
\newblock \href
  {https://pdfs.semanticscholar.org/0cb4/8d8068e5a561c03a070e6de2edc39e4a5437.pdf?_ga=2.252273676.462571563.1612027943-1016261530.1564450085}
  {Evaluating online labor markets for experimental research: Amazon.com's
  {M}echanical {T}urk}.
\newblock In \emph{Political Analysis}, volume~20, pages 351--368. Cambridge
  University Press.

\bibitem[{Bernstein et~al.(2010)Bernstein, Little, Miller, Hartmann, Ackerman,
  Karger, Crowell, and Panovich}]{bernstein2010soylent}
Michael Bernstein, Greg Little, Robert Miller, Björn Hartmann, Mark Ackerman,
  David Karger, David Crowell, and Katrina Panovich. 2010.
\newblock \href {https://doi.org/10.1145/1866029.1866078} {Soylent: A word
  processor with a crowd inside}.
\newblock In \emph{UIST 2010 - 23rd ACM Symposium on User Interface Software
  and Technology}, volume~58, pages 313--322.

\bibitem[{Bie{\'n} et~al.(2020)Bie{\'n}, Gilski, Maciejewska, Taisner,
  Wisniewski, and Lawrynowicz}]{bien-etal-2020-recipenlg}
Micha{\l} Bie{\'n}, Micha{\l} Gilski, Martyna Maciejewska, Wojciech Taisner,
  Dawid Wisniewski, and Agnieszka Lawrynowicz. 2020.
\newblock \href {https://www.aclweb.org/anthology/2020.inlg-1.4}
  {{R}ecipe{NLG}: A cooking recipes dataset for semi-structured text
  generation}.
\newblock In \emph{Proceedings of the 13th International Conference on Natural
  Language Generation}, pages 22--28, Dublin, Ireland. Association for
  Computational Linguistics.

\bibitem[{Brown et~al.(2020)Brown, Mann, Ryder, Subbiah, Kaplan, Dhariwal,
  Neelakantan, Shyam, Sastry, Askell, Agarwal, Herbert-Voss, Krueger, Henighan,
  Child, Ramesh, Ziegler, Wu, Winter, Hesse, Chen, Sigler, Litwin, Gray, Chess,
  Clark, Berner, McCandlish, Radford, Sutskever, and Amodei}]{gpt3}
Tom Brown, Benjamin Mann, Nick Ryder, Melanie Subbiah, Jared~D Kaplan, Prafulla
  Dhariwal, Arvind Neelakantan, Pranav Shyam, Girish Sastry, Amanda Askell,
  Sandhini Agarwal, Ariel Herbert-Voss, Gretchen Krueger, Tom Henighan, Rewon
  Child, Aditya Ramesh, Daniel Ziegler, Jeffrey Wu, Clemens Winter, Chris
  Hesse, Mark Chen, Eric Sigler, Mateusz Litwin, Scott Gray, Benjamin Chess,
  Jack Clark, Christopher Berner, Sam McCandlish, Alec Radford, Ilya Sutskever,
  and Dario Amodei. 2020.
\newblock \href
  {https://proceedings.neurips.cc/paper/2020/file/1457c0d6bfcb4967418bfb8ac142f64a-Paper.pdf}
  {Language models are few-shot learners}.
\newblock In \emph{Advances in Neural Information Processing Systems},
  volume~33, pages 1877--1901. Curran Associates, Inc.

\bibitem[{Callison-Burch et~al.(2007)Callison-Burch, Fordyce, Koehn, Monz, and
  Schroeder}]{callison-burch-etal-2007-meta}
Chris Callison-Burch, Cameron Fordyce, Philipp Koehn, Christof Monz, and Josh
  Schroeder. 2007.
\newblock \href {https://www.aclweb.org/anthology/W07-0718} {(meta-) evaluation
  of machine translation}.
\newblock In \emph{Proceedings of the Second Workshop on Statistical Machine
  Translation}, pages 136--158, Prague, Czech Republic. Association for
  Computational Linguistics.

\bibitem[{Card et~al.(2020)Card, Henderson, Khandelwal, Jia, Mahowald, and
  Jurafsky}]{card-etal-2020-little}
Dallas Card, Peter Henderson, Urvashi Khandelwal, Robin Jia, Kyle Mahowald, and
  Dan Jurafsky. 2020.
\newblock \href {https://doi.org/10.18653/v1/2020.emnlp-main.745} {With little
  power comes great responsibility}.
\newblock In \emph{Proceedings of the 2020 Conference on Empirical Methods in
  Natural Language Processing (EMNLP)}, pages 9263--9274, Online. Association
  for Computational Linguistics.

\bibitem[{Clark and Smith(2021)}]{clark-smith-2021-choose}
Elizabeth Clark and Noah~A. Smith. 2021.
\newblock \href {https://www.aclweb.org/anthology/2021.naacl-main.279} {Choose
  your own adventure: Paired suggestions in collaborative writing for
  evaluating story generation models}.
\newblock In \emph{Proceedings of the 2021 Conference of the North American
  Chapter of the Association for Computational Linguistics: Human Language
  Technologies}, pages 3566--3575, Online. Association for Computational
  Linguistics.

\bibitem[{Daniel et~al.(2018)Daniel, Kucherbaev, Cappiello, Benatallah, and
  Allahbakhsh}]{florian_crowdsourcing}
Florian Daniel, Pavel Kucherbaev, Cinzia Cappiello, Boualem Benatallah, and
  Mohammad Allahbakhsh. 2018.
\newblock \href {https://doi.org/10.1145/3148148} {Quality control in
  crowdsourcing: A survey of quality attributes, assessment techniques, and
  assurance actions}.
\newblock In \emph{ACM Computing Surveys}, volume~51. Association for Computing
  Machinery.

\bibitem[{Dugan et~al.(2020)Dugan, Ippolito, Kirubarajan, and
  Callison-Burch}]{dugan-etal-2020-roft}
Liam Dugan, Daphne Ippolito, Arun Kirubarajan, and Chris Callison-Burch. 2020.
\newblock \href {https://doi.org/10.18653/v1/2020.emnlp-demos.25} {{R}o{FT}: A
  tool for evaluating human detection of machine-generated text}.
\newblock In \emph{Proceedings of the 2020 Conference on Empirical Methods in
  Natural Language Processing: System Demonstrations}, pages 189--196, Online.
  Association for Computational Linguistics.

\bibitem[{Fan et~al.(2018)Fan, Lewis, and Dauphin}]{fan-etal-2018-hierarchical}
Angela Fan, Mike Lewis, and Yann Dauphin. 2018.
\newblock \href {https://doi.org/10.18653/v1/P18-1082} {Hierarchical neural
  story generation}.
\newblock In \emph{Proceedings of the 56th Annual Meeting of the Association
  for Computational Linguistics (Volume 1: Long Papers)}, pages 889--898,
  Melbourne, Australia. Association for Computational Linguistics.

\bibitem[{Garbacea et~al.(2019)Garbacea, Carton, Yan, and
  Mei}]{garbacea-etal-2019-judge}
Cristina Garbacea, Samuel Carton, Shiyan Yan, and Qiaozhu Mei. 2019.
\newblock \href {https://doi.org/10.18653/v1/D19-1409} {Judge the judges: A
  large-scale evaluation study of neural language models for online review
  generation}.
\newblock In \emph{Proceedings of the 2019 Conference on Empirical Methods in
  Natural Language Processing and the 9th International Joint Conference on
  Natural Language Processing (EMNLP-IJCNLP)}, pages 3968--3981, Hong Kong,
  China. Association for Computational Linguistics.

\bibitem[{Gehrmann et~al.(2021)Gehrmann, Adewumi, Aggarwal, Ammanamanchi,
  Anuoluwapo, Bosselut, Chandu, Clinciu, Das, Dhole, Du, Durmus, Dušek,
  Emezue, Gangal, Garbacea, Hashimoto, Hou, Jernite, Jhamtani, Ji, Jolly, Kale,
  Kumar, Ladhak, Madaan, Maddela, Mahajan, Mahamood, Majumder, Martins,
  McMillan-Major, Mille, van Miltenburg, Nadeem, Narayan, Nikolaev, Niyongabo,
  Osei, Parikh, Perez-Beltrachini, Rao, Raunak, Rodriguez, Santhanam, Sedoc,
  Sellam, Shaikh, Shimorina, Cabezudo, Strobelt, Subramani, Xu, Yang, Yerukola,
  and Zhou}]{gem}
Sebastian Gehrmann, Tosin Adewumi, Karmanya Aggarwal, Pawan~Sasanka
  Ammanamanchi, Aremu Anuoluwapo, Antoine Bosselut, Khyathi~Raghavi Chandu,
  Miruna Clinciu, Dipanjan Das, Kaustubh~D. Dhole, Wanyu Du, Esin Durmus,
  Ondřej Dušek, Chris Emezue, Varun Gangal, Cristina Garbacea, Tatsunori
  Hashimoto, Yufang Hou, Yacine Jernite, Harsh Jhamtani, Yangfeng Ji, Shailza
  Jolly, Mihir Kale, Dhruv Kumar, Faisal Ladhak, Aman Madaan, Mounica Maddela,
  Khyati Mahajan, Saad Mahamood, Bodhisattwa~Prasad Majumder, Pedro~Henrique
  Martins, Angelina McMillan-Major, Simon Mille, Emiel van Miltenburg, Moin
  Nadeem, Shashi Narayan, Vitaly Nikolaev, Rubungo~Andre Niyongabo, Salomey
  Osei, Ankur Parikh, Laura Perez-Beltrachini, Niranjan~Ramesh Rao, Vikas
  Raunak, Juan~Diego Rodriguez, Sashank Santhanam, João Sedoc, Thibault
  Sellam, Samira Shaikh, Anastasia Shimorina, Marco Antonio~Sobrevilla
  Cabezudo, Hendrik Strobelt, Nishant Subramani, Wei Xu, Diyi Yang, Akhila
  Yerukola, and Jiawei Zhou. 2021.
\newblock \href {https://arxiv.org/abs/2102.01672} {The {GEM} benchmark:
  Natural language generation, its evaluation and metrics}.
\newblock \emph{ArXiv}, abs/2102.01672.

\bibitem[{Gehrmann et~al.(2019)Gehrmann, Strobelt, and
  Rush}]{gehrmann-etal-2019-gltr}
Sebastian Gehrmann, Hendrik Strobelt, and Alexander Rush. 2019.
\newblock \href {https://doi.org/10.18653/v1/P19-3019} {{GLTR}: Statistical
  detection and visualization of generated text}.
\newblock In \emph{Proceedings of the 57th Annual Meeting of the Association
  for Computational Linguistics: System Demonstrations}, pages 111--116,
  Florence, Italy. Association for Computational Linguistics.

\bibitem[{Hashimoto et~al.(2019)Hashimoto, Zhang, and
  Liang}]{hashimoto-etal-2019-unifying}
Tatsunori Hashimoto, Hugh Zhang, and Percy Liang. 2019.
\newblock \href {https://doi.org/10.18653/v1/N19-1169} {Unifying human and
  statistical evaluation for natural language generation}.
\newblock In \emph{Proceedings of the 2019 Conference of the North {A}merican
  Chapter of the Association for Computational Linguistics: Human Language
  Technologies, Volume 1 (Long and Short Papers)}, pages 1689--1701,
  Minneapolis, Minnesota. Association for Computational Linguistics.

\bibitem[{Howcroft et~al.(2020)Howcroft, Belz, Clinciu, Gkatzia, Hasan,
  Mahamood, Mille, van Miltenburg, Santhanam, and
  Rieser}]{howcroft-etal-2020-twenty}
David~M. Howcroft, Anya Belz, Miruna-Adriana Clinciu, Dimitra Gkatzia, Sadid~A.
  Hasan, Saad Mahamood, Simon Mille, Emiel van Miltenburg, Sashank Santhanam,
  and Verena Rieser. 2020.
\newblock \href {https://www.aclweb.org/anthology/2020.inlg-1.23} {Twenty years
  of confusion in human evaluation: {NLG} needs evaluation sheets and
  standardised definitions}.
\newblock In \emph{Proceedings of the 13th International Conference on Natural
  Language Generation}, pages 169--182, Dublin, Ireland. Association for
  Computational Linguistics.

\bibitem[{Ippolito et~al.(2020)Ippolito, Duckworth, Callison-Burch, and
  Eck}]{ippolito-etal-2020-automatic}
Daphne Ippolito, Daniel Duckworth, Chris Callison-Burch, and Douglas Eck. 2020.
\newblock \href {https://doi.org/10.18653/v1/2020.acl-main.164} {Automatic
  detection of generated text is easiest when humans are fooled}.
\newblock In \emph{Proceedings of the 58th Annual Meeting of the Association
  for Computational Linguistics}, pages 1808--1822, Online. Association for
  Computational Linguistics.

\bibitem[{Khashabi et~al.(2021)Khashabi, Stanovsky, Bragg, Lourie, Kasai, Choi,
  Smith, and Weld}]{genie}
Daniel Khashabi, Gabriel Stanovsky, Jonathan Bragg, Nicholas Lourie, Jungo
  Kasai, Yejin Choi, Noah~A. Smith, and Daniel~S. Weld. 2021.
\newblock \href {https://arxiv.org/abs/2101.06561} {{GENIE}: A leaderboard for
  human-in-the-loop evaluation of text generation}.
\newblock \emph{ArXiv}, abs/2101.06561.

\bibitem[{Kim et~al.(2017)Kim, Sterman, Cohen, and
  Bernstein}]{kim2017mechanical}
Joy Kim, Sarah Sterman, Allegra Argent~Beal Cohen, and Michael~S Bernstein.
  2017.
\newblock \href {https://dl.acm.org/doi/10.1145/2998181.2998196} {Mechanical
  novel: Crowdsourcing complex work through reflection and revision}.
\newblock In \emph{Proceedings of the 2017 ACM Conference on Computer Supported
  Cooperative Work and Social Computing}, pages 233--245. Association for
  Computing Machinery.

\bibitem[{Kittur and Kraut(2008)}]{10.1145/1460563.1460572}
Aniket Kittur and Robert~E. Kraut. 2008.
\newblock \href {https://doi.org/10.1145/1460563.1460572} {Harnessing the
  wisdom of crowds in wikipedia: Quality through coordination}.
\newblock In \emph{Proceedings of the 2008 ACM Conference on Computer Supported
  Cooperative Work}, CSCW '08, page 37–46, New York, NY, USA. Association for
  Computing Machinery.

\bibitem[{van~der Lee et~al.(2021)van~der Lee, Gatt, van Miltenburg, and
  Krahmer}]{vanderlee_journal}
Chris van~der Lee, Albert Gatt, Emiel van Miltenburg, and Emiel Krahmer. 2021.
\newblock \href {https://doi.org/https://doi.org/10.1016/j.csl.2020.101151}
  {Human evaluation of automatically generated text: Current trends and best
  practice guidelines}.
\newblock \emph{Computer Speech \& Language}, 67:101151.

\bibitem[{van~der Lee et~al.(2019)van~der Lee, Gatt, van Miltenburg, Wubben,
  and Krahmer}]{van-der-lee-etal-2019-best}
Chris van~der Lee, Albert Gatt, Emiel van Miltenburg, Sander Wubben, and Emiel
  Krahmer. 2019.
\newblock \href {https://doi.org/10.18653/v1/W19-8643} {Best practices for the
  human evaluation of automatically generated text}.
\newblock In \emph{Proceedings of the 12th International Conference on Natural
  Language Generation}, pages 355--368, Tokyo, Japan. Association for
  Computational Linguistics.

\bibitem[{Liu et~al.(2016)Liu, Lowe, Serban, Noseworthy, Charlin, and
  Pineau}]{liu-etal-2016-evaluate}
Chia-Wei Liu, Ryan Lowe, Iulian Serban, Mike Noseworthy, Laurent Charlin, and
  Joelle Pineau. 2016.
\newblock \href {https://doi.org/10.18653/v1/D16-1230} {How {NOT} to evaluate
  your dialogue system: An empirical study of unsupervised evaluation metrics
  for dialogue response generation}.
\newblock In \emph{Proceedings of the 2016 Conference on Empirical Methods in
  Natural Language Processing}, pages 2122--2132, Austin, Texas. Association
  for Computational Linguistics.

\bibitem[{Mitra et~al.(2015)Mitra, Hutto, and Gilbert}]{mitra_crowdsourcing}
Tanushree Mitra, C.J. Hutto, and Eric Gilbert. 2015.
\newblock \href {https://dl.acm.org/doi/10.1145/2702123.2702553} {Comparing
  person- and process-centric strategies for obtaining quality data on {A}mazon
  {M}echanical {T}urk}.
\newblock In \emph{Proceedings of the 33rd Annual ACM Conference on Human
  Factors in Computing Systems}, pages 1345--1354. Association for Computing
  Machinery.

\bibitem[{Novikova et~al.(2017)Novikova, Du{\v{s}}ek, Cercas~Curry, and
  Rieser}]{novikova-etal-2017-need}
Jekaterina Novikova, Ond{\v{r}}ej Du{\v{s}}ek, Amanda Cercas~Curry, and Verena
  Rieser. 2017.
\newblock \href {https://doi.org/10.18653/v1/D17-1238} {Why we need new
  evaluation metrics for {NLG}}.
\newblock In \emph{Proceedings of the 2017 Conference on Empirical Methods in
  Natural Language Processing}, pages 2241--2252, Copenhagen, Denmark.
  Association for Computational Linguistics.

\bibitem[{Novikova et~al.(2018)Novikova, Du{\v{s}}ek, and
  Rieser}]{novikova-etal-2018-rankme}
Jekaterina Novikova, Ond{\v{r}}ej Du{\v{s}}ek, and Verena Rieser. 2018.
\newblock \href {https://doi.org/10.18653/v1/N18-2012} {{R}ank{ME}: Reliable
  human ratings for natural language generation}.
\newblock In \emph{Proceedings of the 2018 Conference of the North {A}merican
  Chapter of the Association for Computational Linguistics: Human Language
  Technologies, Volume 2 (Short Papers)}, pages 72--78, New Orleans, Louisiana.
  Association for Computational Linguistics.

\bibitem[{Oppenheimer et~al.(2009)Oppenheimer, Meyvis, and
  Davidenko}]{oppenheimer_crowdsourcing}
Daniel~M. Oppenheimer, Tom Meyvis, and Nicolas Davidenko. 2009.
\newblock \href {https://doi.org/https://doi.org/10.1016/j.jesp.2009.03.009}
  {Instructional manipulation checks: Detecting satisficing to increase
  statistical power}.
\newblock In \emph{Journal of Experimental Social Psychology}, volume~45, pages
  867--872. Elsevier.

\bibitem[{Radford et~al.(2019)Radford, Wu, Child, Luan, Amodei, and
  Sutskever}]{gpt2}
Alec Radford, Jeff Wu, Rewon Child, David Luan, Dario Amodei, and Ilya
  Sutskever. 2019.
\newblock \href
  {https://d4mucfpksywv.cloudfront.net/better-language-models/language-models.pdf}
  {Language models are unsupervised multitask learners}.
\newblock \emph{OpenAI Blog}.

\bibitem[{Santhanam and Shaikh(2019)}]{santhanam-shaikh-2019-towards}
Sashank Santhanam and Samira Shaikh. 2019.
\newblock \href {https://doi.org/10.18653/v1/W19-8610} {Towards best experiment
  design for evaluating dialogue system output}.
\newblock In \emph{Proceedings of the 12th International Conference on Natural
  Language Generation}, pages 88--94, Tokyo, Japan. Association for
  Computational Linguistics.

\bibitem[{Schoch et~al.(2020)Schoch, Yang, and Ji}]{schoch-etal-2020-problem}
Stephanie Schoch, Diyi Yang, and Yangfeng Ji. 2020.
\newblock \href {https://www.aclweb.org/anthology/2020.evalnlgeval-1.2}
  {{``}{T}his is a problem, don{'}t you agree?{''} {F}raming and bias in human
  evaluation for natural language generation}.
\newblock In \emph{Proceedings of the 1st Workshop on Evaluating NLG
  Evaluation}, pages 10--16, Online (Dublin, Ireland). Association for
  Computational Linguistics.

\bibitem[{Turing(1950)}]{turing_test}
Alan Turing. 1950.
\newblock \href {https://doi.org/10.1093/mind/lix.236.433} {{Computing}
  {Machinery} {and} {Intelligence}}.
\newblock In \emph{Mind}, volume {LIX}, pages 433--460. Oxford University Press
  ({OUP}).

\bibitem[{Uchendu et~al.(2020)Uchendu, Le, Shu, and
  Lee}]{uchendu-etal-2020-authorship}
Adaku Uchendu, Thai Le, Kai Shu, and Dongwon Lee. 2020.
\newblock \href {https://doi.org/10.18653/v1/2020.emnlp-main.673} {Authorship
  attribution for neural text generation}.
\newblock In \emph{Proceedings of the 2020 Conference on Empirical Methods in
  Natural Language Processing (EMNLP)}, pages 8384--8395, Online. Association
  for Computational Linguistics.

\bibitem[{Weld et~al.(2014)Weld, Mausam, Lin, and Bragg}]{weld_crowdsourcing}
Daniel~S. Weld, Mausam, Christopher~H. Lin, and Jonathan Bragg. 2014.
\newblock \href
  {https://homes.cs.washington.edu/~weld/papers/ci-chapter2014.pdf} {Artificial
  intelligence and collective intelligence}.
\newblock In \emph{Handbook of Collective Intelligence}, chapter~3. MIT Press.

\bibitem[{Zellers et~al.(2019{\natexlab{a}})Zellers, Holtzman, Bisk, Farhadi,
  and Choi}]{zellers-etal-2019-hellaswag}
Rowan Zellers, Ari Holtzman, Yonatan Bisk, Ali Farhadi, and Yejin Choi.
  2019{\natexlab{a}}.
\newblock \href {https://doi.org/10.18653/v1/P19-1472} {{H}ella{S}wag: Can a
  machine really finish your sentence?}
\newblock In \emph{Proceedings of the 57th Annual Meeting of the Association
  for Computational Linguistics}, pages 4791--4800, Florence, Italy.
  Association for Computational Linguistics.

\bibitem[{Zellers et~al.(2021)Zellers, Holtzman, Clark, Qin, Farhadi, and
  Choi}]{zellers_turingadvice}
Rowan Zellers, Ari Holtzman, Elizabeth Clark, Lianhui Qin, Ali Farhadi, and
  Yejin Choi. 2021.
\newblock \href {https://www.aclweb.org/anthology/2021.naacl-main.386}
  {{T}uring{A}dvice: A generative and dynamic evaluation of language use}.
\newblock In \emph{Proceedings of the 2021 Conference of the North American
  Chapter of the Association for Computational Linguistics: Human Language
  Technologies}, pages 4856--4880, Online. Association for Computational
  Linguistics.

\bibitem[{Zellers et~al.(2019{\natexlab{b}})Zellers, Holtzman, Rashkin, Bisk,
  Farhadi, Roesner, and Choi}]{grover_zellers}
Rowan Zellers, Ari Holtzman, Hannah Rashkin, Yonatan Bisk, Ali Farhadi,
  Franziska Roesner, and Yejin Choi. 2019{\natexlab{b}}.
\newblock \href
  {https://proceedings.neurips.cc/paper/2019/file/3e9f0fc9b2f89e043bc6233994dfcf76-Paper.pdf}
  {Defending against neural fake news}.
\newblock In \emph{Advances in Neural Information Processing Systems},
  volume~32. Curran Associates, Inc.

\bibitem[{Zhang et~al.(2020)Zhang, Zhao, Saleh, and Liu}]{Zhang2020PEGASUSPW}
Jingqing Zhang, Yao Zhao, Mohammad Saleh, and Peter Liu. 2020.
\newblock \href {http://proceedings.mlr.press/v119/zhang20ae.html} {{PEGASUS}:
  Pre-training with extracted gap-sentences for abstractive summarization}.
\newblock In \emph{Proceedings of the 37th International Conference on Machine
  Learning}, volume 119, pages 11328--11339. PMLR.

\end{thebibliography}

\appendix
\clearpage\newpage
\appendix
\section{Appendices}
\label{sec:appendix}

\subsection{Newspapers}\label{app:newspapers}
Each newspaper came from a randomly chosen U.S. state and was selected from Wikipedia's lists of newspapers by state (\url{en.wikipedia.org/wiki/List_of_newspapers_in_the_United_States\#By_state_and_territory}).
The human-authored news articles and prompts came from the following states and websites:
\begin{itemize}[noitemsep]
    \item HI: \url{www.westhawaiitoday.com}
    \item CT: \url{www.greenwichtime.com/}
    \item WA: \url{www.vashonbeachcomber.com/}
    \item SD: \url{www.argusleader.com/}
    \item CA: \url{www.redding.com/}
    \item MA: \url{www.lowellsun.com/}
    \item NE: \url{starherald.com/}
    \item VA: \url{dailyprogress.com/}
    \item WV: \url{www.theintermountain.com/}
    \item NM: \url{www.lcsun-news.com/}
    \item LA: \url{www.nola.com/}
    \item IA: \url{qctimes.com/}
    \item NY: \url{www.pressconnects.com/}
    \item IN: \url{www.pal-item.com/}
    \item NJ: \url{www.northjersey.com/}
\end{itemize}

\subsection{Score Frequencies}\label{app:exp1_histograms}
\begin{figure*}[h]
     \centering
     \begin{subfigure}[b]{0.24\textwidth}
         \centering
         \includegraphics[width=\textwidth]{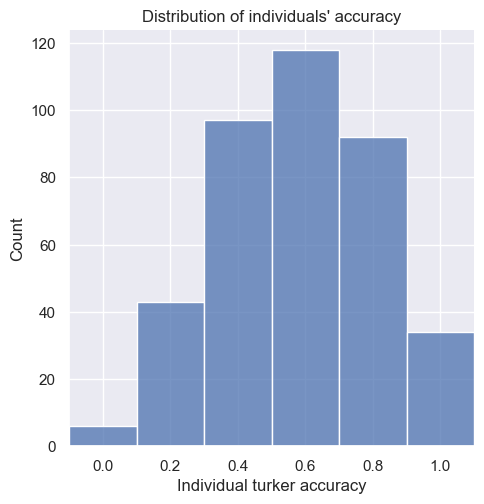}
         \caption{GPT2 overall}
         \label{fig:hist_gpt2_all}
     \end{subfigure}
     \hfill
     \begin{subfigure}[b]{0.24\textwidth}
         \centering
         \includegraphics[width=\textwidth]{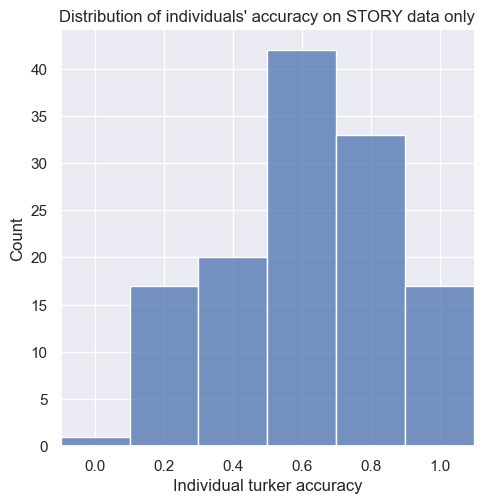}
         \caption{GPT2 story}
         \label{fig:hist_gpt2_story}
     \end{subfigure}
     \hfill
     \begin{subfigure}[b]{0.24\textwidth}
         \centering
         \includegraphics[width=\textwidth]{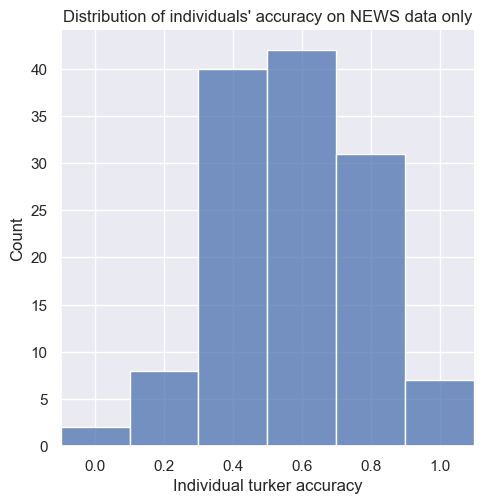}
         \caption{GPT2 news}
         \label{fig:hist_gpt2_news}
     \end{subfigure}
      \hfill
     \begin{subfigure}[b]{0.24\textwidth}
         \centering
         \includegraphics[width=\textwidth]{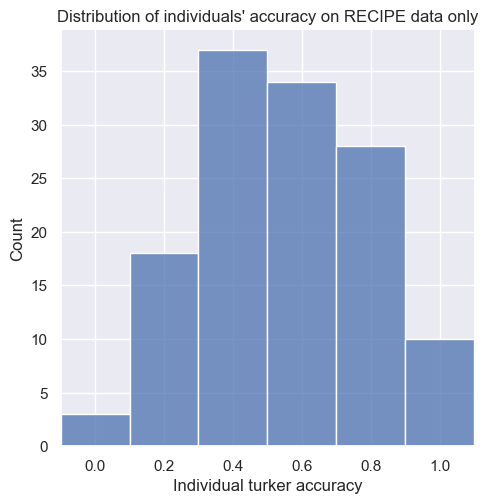}
         \caption{GPT2 recipe}
         \label{fig:hist_gpt2_recipe}
     \end{subfigure}
        \caption{Histogram of scores classifying human and GPT2 texts.}
        \label{fig:gpt2_histograms}
\end{figure*}

\begin{figure*}[h]
     \centering
     \begin{subfigure}[b]{0.24\textwidth}
         \centering
         \includegraphics[width=\textwidth]{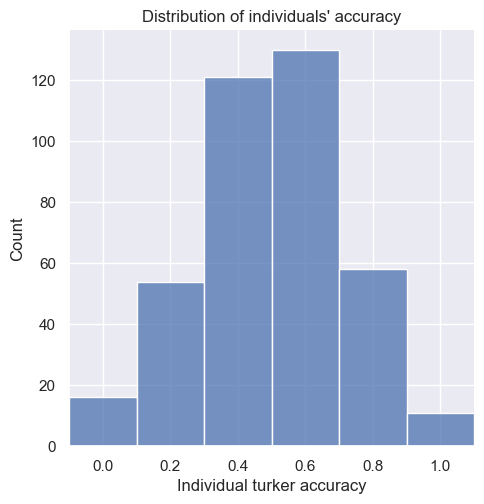}
         \caption{GPT3 overall}
         \label{fig:hist_gpt3_all}
     \end{subfigure}
     \hfill
     \begin{subfigure}[b]{0.24\textwidth}
         \centering
         \includegraphics[width=\textwidth]{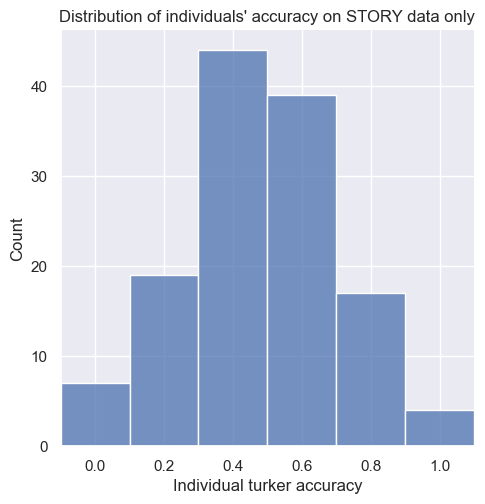}
         \caption{GPT3 story}
         \label{fig:hist_gpt3_story}
     \end{subfigure}
     \hfill
     \begin{subfigure}[b]{0.24\textwidth}
         \centering
         \includegraphics[width=\textwidth]{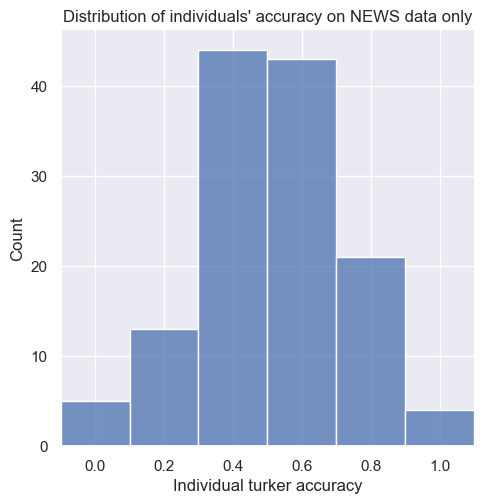}
         \caption{GPT3 news}
         \label{fig:hist_gpt3_news}
     \end{subfigure}
      \hfill
     \begin{subfigure}[b]{0.24\textwidth}
         \centering
         \includegraphics[width=\textwidth]{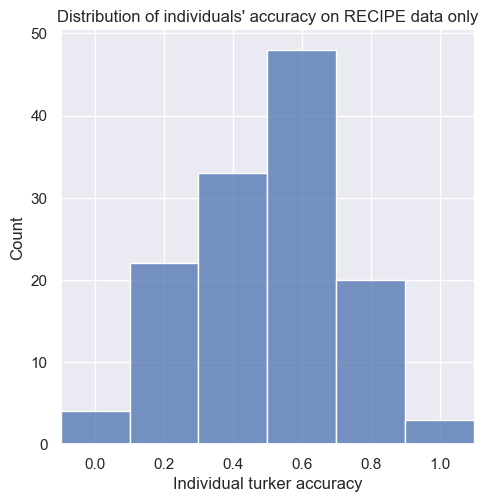}
         \caption{GPT3 recipe}
         \label{fig:hist_gpt3_recipe}
     \end{subfigure}
        \caption{Histogram of scores classifying human and GPT3 texts.}
        \label{fig:gpt3_histograms}
\end{figure*}

The frequency of the scores (out of 5) received by evaluators is shown in Figures \ref{fig:gpt2_histograms} (for GPT2 experiments) and \ref{fig:gpt3_histograms} (for GPT3 experiments).

\subsection{Annotation Details}\label{app:annotation}
The authors annotated 300 comments (150 from the No Training experiment and 150 from the Examples experiment).
For each experiment, we randomly chose 50 authors from each setting and randomly added 1 of their responses to the annotation set.
Each comment was annotated by 2 of the authors.
The annotation labels are shown in Table \ref{tab:annotation_labels}.
To create the set of annotation labels, the authors created a candidate list of labels, annotated a subset of the data collected in the pilot study (Appendix \ref{app:pilot}) together, then another subset separately, and finally refined the labels based on feedback from that process.
Because evaluators' responses often contained more than one reason for their choice, comments could receive more than one label.

\subsection{Evaluators' Expectations of Generated Text}\label{app:HUM}
Because we asked evaluators whether they thought the text was human- or machine-authored, they often justified their choices by explaining what types of human language they believed machines could (or could not) generate.
We took note of these comments and annotated for them in our data annotation process (Appendix \ref{app:annotation}) because they demonstrate the expectations evaluators have for the quality of machine-generated text.
Some example comments shown in Table \ref{tab:HUM}.

\begin{table*}[t!]
\centering
\small
\begin{tabular}{p{.09\textwidth}p{0.1\textwidth}p{.3\textwidth}p{.4\textwidth}}
\toprule
\textbf{Category} & \textbf{Label} & \textbf{Description} & \textbf{Example} \\
\midrule
\multirow{3}{*}{Form} & Grammar & The spelling and grammar of the text, punctuation/formatting issues & I would make the text more grammatical by adding more punctuation where necassary. \\
 & Level of detail & Is the text simple or does it go more in-depth? & i would include more examples and explanations of the statements. The author need to elaborate more on the topic. \\
 & Genre & If the text is the genre/domain/style/formality that the reader expects, adheres to style norms & written exactly the way a human will tell a story \\
 \midrule
\multirow{5}{*}{Content} & Repetition & Words/phrases/content repeated itself & Repeating ``or some would say'' seemed very unnatural. \\
 & Factuality & The accuracy of the text, whether it describes things that are ``true.'' & The article lists many facts that make the information seem like it was machine-generated. \\
 & Consistency & How the text relates to the context and other pieces of the text & The subject of the article follows the headline well without repeating it exactly \\
 & Common sense & Whether the text ``makes sense'' within the world that it is written & Change the ``bake in the preheated oven for 20 minutes on top of the stove.'' You can't bake on top of the stove but to bake in the oven. \\
 & Coherence & The structure and coherence of the text. Order issues go here. & More cohesion between sentences. Feel loosely related, but wording is strange. \\
 \midrule
Machine capabilities & Writer intent and expression & Speculating about writer's intent or capabilities (e.g., ability to express emotions) & The text is thorough and tries to cover all basis of the situation. It is very inclusive and humans worry about being inclusive not machines. \\
\midrule
\multirow{2}{*}{Null} & Miscellaneous & Everything else & too many dialogue-like things, and make it less gender-dicey. \\
 & Null/Vague & No reasons given, or too vague to be considered a real reason & i selected this rating because it is definitely written by human \\
 \bottomrule
\end{tabular}
\caption{The annotation labels, along with an example of each label. Note that some example sentences would also be labeled with additional labels. We did not use the Null category in the paper's analyses.}
\label{tab:annotation_labels}
\end{table*}

\begin{table*}[t!]
\centering
\small
\begin{tabular}{p{\linewidth}}
\toprule
Punctuation is perfect as well as the flow of the text. There is also more complex punctuation, such as quotes, that I think a computer would get wrong. \\
\midrule
``fried anyone to a crisp.'' That is a human if I've ever seen one. a bot or AI is more proper, they wouldn't write so casual. \\
\midrule
Because it talked about love which robots know nothing about. \\
\midrule
Lack of oxford comma. A computer would know better. \\
\midrule
The article flows properly, has appropriate English and multiple quotes. This would seem to be more than a bot could create. How would a bot create random quotes? \\
\midrule
This was more of a ramble which humans do, not computers. \\
\midrule
There are details and key phrases used in this article that computer generated text would not have in it, such as ``came up short'', ``put together a solid drive'', ``put up any points''. These are human specific terms and are not generally able to be programmed into a text program. \\
\midrule
This piece quotes the host and I don't believe AI can interview people yet so this has to be human written. \\
\midrule
It has a lot of detail in an emotional description that a machine isn't capable of giving to its readers. \\
\midrule
The way some words are phrased here again shows the human uncertainty, ``let the apples marinate for about 30 minutes''. If this was machine-generated, it would most likely just say marinate for 30 minutes. \\
\midrule
It seems to know when to use semicolns very well. This could be a human or a really smart computer. \\
\midrule
I don’t think AIs are capable of writing recipes on their own just yet. \\
\midrule
I don't believe a machine could come up with this level of whimsy or creativity and have it make sense. \\
\midrule
I don't think AI would use the term `literally'. \\
\midrule
There is a lot of every day language written in this recipe that I couldn't see a machine possibly replicating. \\
\midrule
It adds that she is both nervous and excited whereas a machine wouldn't care what emotions are involved. \\
\midrule
The writer used proper grammar and punctuation. No bot could write this, \\
\midrule
I'm not sure if a computer would get the concept or use the word ``your'' where the recipe begins with ``Start by doing your prep.'' \\
\bottomrule
\end{tabular}
\caption{Example reasons evaluators gave for their decisions that spoke to their beliefs about current NLG capabilities.}
\label{tab:HUM}
\end{table*}

\subsection{Pilot Study}\label{app:pilot}
Before running the experiments described in the paper, we ran a smaller-scale version with both Amazon Mechanical Turk ($n=22$) and ``expert'' evaluators (NLP graduate students; $n=11$).
We asked the evaluators to distinguish between stories authored by humans, GPT2, and GPT3 and to explain their reasoning.
When we coded and analyzed their responses, we found that the most accurate evaluators focused on textual aspects like repetition and were less likely to mention aspects like style.
The AMT evaluators mentioned grammar and spelling far more frequently than the expert evaluators, who were more likely to mention the repetition, factuality, and commonsense of the passage.

\subsection{Training and Instructions}\label{app:training+instructions}
\begin{figure*}
\centering
\fbox{\includegraphics[scale=0.4, angle=270, trim={6.5cm 1cm 8cm 0.75cm},clip]{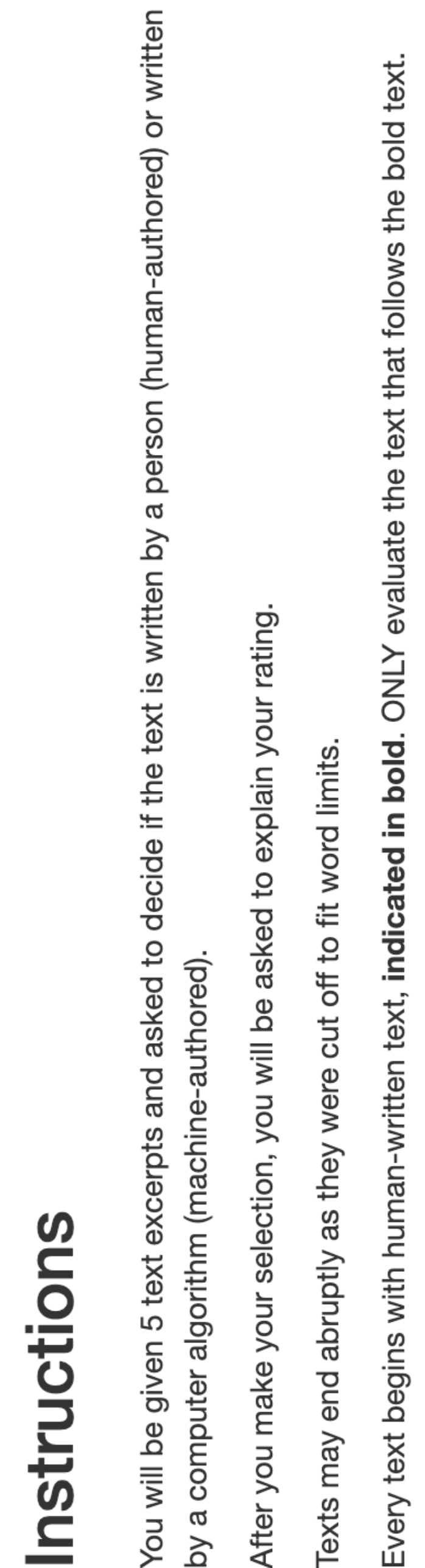}}
\caption{Basic instructions shown to all evaluators.}
\label{fig:basic_instr}
\end{figure*}

Figure \ref{fig:basic_instr} shows the basic instructions that were shown to all evaluators, in both \S\ref{sec:exp_1} and \S\ref{sec:exp_2}, regardless of training or domain.
All training information occurred after receiving the basic instructions.

\subsubsection{Instruction Training}\label{app:train_instructions}
\begin{figure*}
\centering
\fbox{\includegraphics[scale=0.4, angle=270, trim={9.5cm 1cm 3.5cm 1cm},clip]{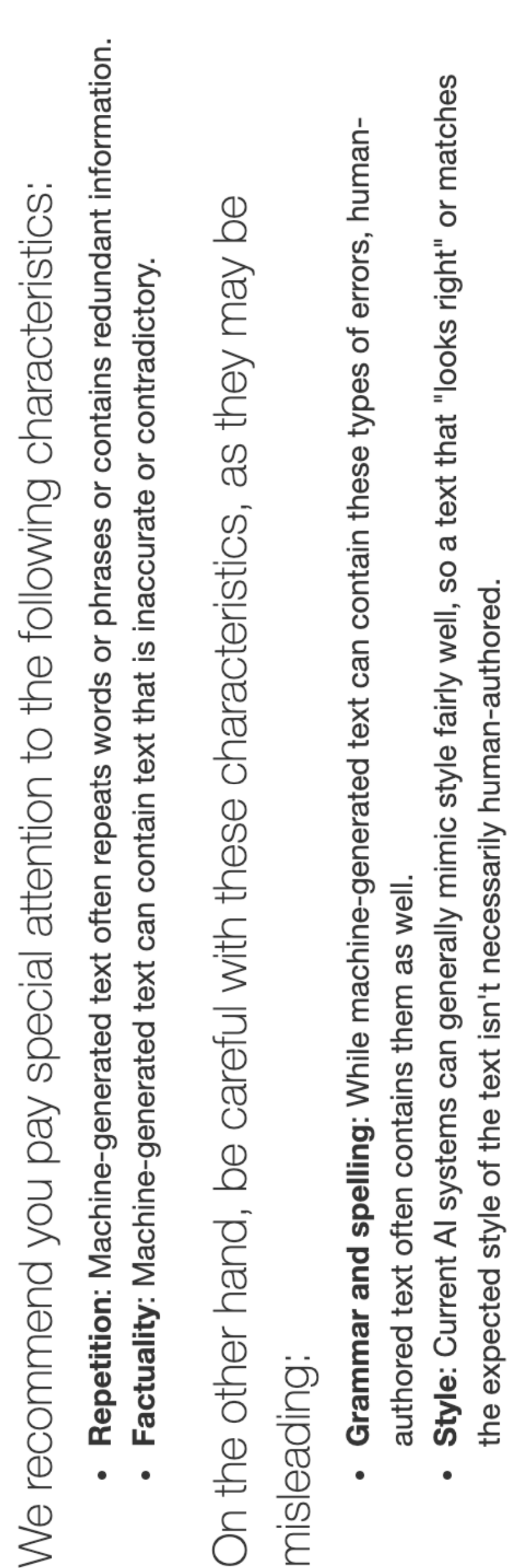}}
\caption{The Instruction training.}
\label{fig:training_instr}
\end{figure*}

The training shown to evaluators in the Instruction training condition is shown in Figure \ref{fig:training_instr}.

\subsubsection{Example Training}\label{app:train_examples}
\begin{figure*}[]
\centering
\includegraphics[scale=0.2, trim={1cm 0cm 1cm 1cm},clip]{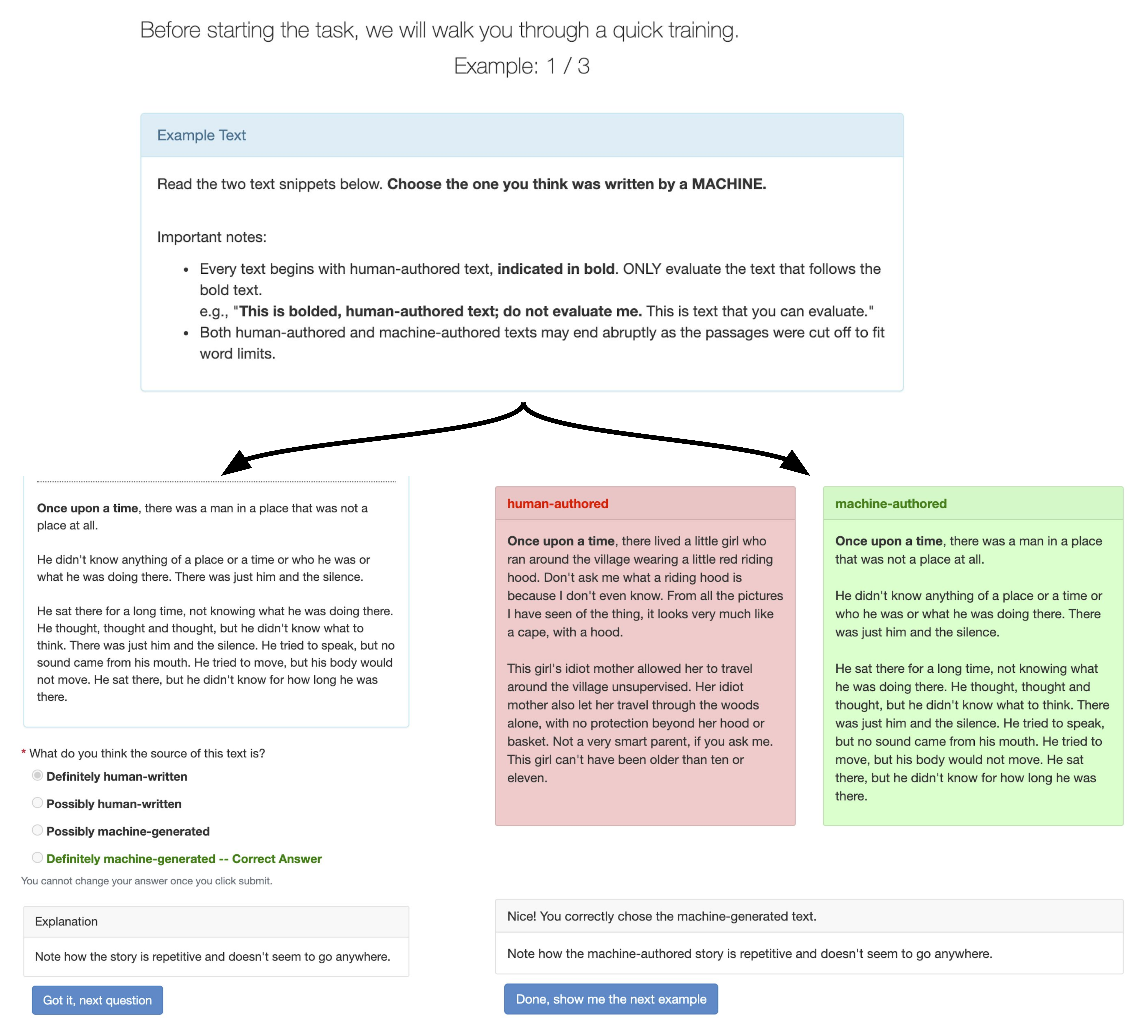}
\caption{The Example training (left) and Comparison training (right) in the story domain. The instructions are the same for both, except ``Choose the one you think was written by a machine.'' was in Comparison only.}
\label{fig:training_example_compare}
\end{figure*}

A screenshot of the Examples and Comparison training is in Figure \ref{fig:training_example_compare}. The full set of examples and annotations used in the Examples and Comparison trainings can be found in the supplementary materials and at \url{ark.cs.washington.edu/human_evals_ACL21}.

\end{document}